\pdfoutput=1

\documentclass[11pt]{article}

\usepackage[preprint]{acl}

\usepackage{times}
\usepackage{latexsym}
\usepackage{arydshln}
\usepackage{tabularray}
\usepackage{multirow}
\usepackage{booktabs, caption, array}
\usepackage{algorithm}
\usepackage{algorithmic}
\usepackage{comment}
\usepackage{xcolor}
\usepackage{amssymb}
\usepackage{subfigure}
\usepackage{graphicx}
\usepackage{enumitem}
\usepackage{framed}
\usepackage{amsmath}
\definecolor{shadecolor}{gray}{0.9}
\usepackage{comment}
\usepackage{xcolor}
\usepackage{hyperref}
 \definecolor{darkblue}{rgb}{0, 0, 0.5}
  \hypersetup{colorlinks=true, citecolor=darkblue, linkcolor=darkblue, urlcolor=darkblue}
\usepackage{graphicx}
\usepackage{tabularx}
\usepackage{soul}
\usepackage[table]{colortbl}
\usepackage{xstring}

\usepackage{color}
\usepackage[T1]{fontenc}

\usepackage[utf8]{inputenc}

\usepackage{microtype}

\usepackage{inconsolata}

\title{Effective In-Context Example Selection through Data Compression}

\author{Zhongxiang Sun\textsuperscript{{$\star$}} \and Kepu Zhang\textsuperscript{{$\star$}} \and Haoyu Wang \and Xiao Zhang \and Jun Xu\textsuperscript{{$\dagger$}} \\
      Gaoling School of Artificial Intelligence,  \\ Renmin University of China \\ 
      \texttt{\{sunzhongxiang, kepuzhang, wanghaoyu0924, zhangx89, junxu\}@ruc.edu.cn}  }

\begin{document}
\maketitle
\let\thefootnote\relax\footnotetext{$^\star$ Equal contribution.\hspace{3pt}}
\let\thefootnote\relax\footnotetext{$^\dagger$ Jun Xu is the corresponding author. \hspace{3pt} }

\begin{abstract}
In-context learning has been extensively validated in large language models. However, the mechanism and selection strategy for in-context example selection, which is a crucial ingredient in this approach, lacks systematic and in-depth research. In this paper, we propose a data compression approach to the selection of in-context examples. We introduce a two-stage method that can effectively choose relevant examples and retain sufficient information about the training dataset within the in-context examples. Our method shows a significant improvement of an average of 5.90\% across five different real-world datasets using four language models.
\end{abstract}

\section{Introduction}
Drawing inspiration from recent research that regards Large Language Models (LLMs) as an efficient means of compressing pre-training datasets, and the notion that In-Context Learning (ICL) can be seen as fine-tuning on example datasets~\citep{dai2022can}, we assume that LLMs can achieve data re-compression through ICL. In other words, an effective training dataset compression method can aid in the selection of in-context examples. Looking at the matter from another perspective, it is evident that fine-tuning the entire dataset would yield the best results. However, in the case of ICL, we typically choose only a few examples as LLM prompts due to the limitations of input window length. By employing data compression techniques, we can ensure that the majority of data information is preserved in the in-context examples, which is also the aim of dataset pruning.

Based on the aforementioned analysis, we propose to utilize the influence function~\cite{influence_function}, which has exhibited efficacy in dataset pruning, to select examples for ICL. However, recent studies on ICL have revealed that the relevance between the query source and in-context examples is critical for ICL. Furthermore, the influence function requires the gradient of parameters, which is computationally expensive and inefficient. To tackle the aforementioned issues, we suggest a two-stage method. Firstly, relevant examples for the query input are recalled, which ensures the correlation between the examples and the query source. Secondly, our meta-gradient-based influence function is utilized to calculate the influence score for each recalled example. Finally, based on the influence score, in-context examples are selected from the recalled examples. Notably, our framework compresses important information from the training set into the in-context examples, thereby enhancing the performance of ICL. Additionally, our framework is data-independent, relies solely on a small number of model parameters, and does not require the training of any additional models. Numerous experiments indicate that our method shows a significant improvement of an average of 5.90\% on five different real-world datasets using multiple language models.

\section{Background}
\subsection{In-Context Learning}
The ICL scenario of LLMs can be regarded as a conditional text generation problem. Concretely, the probability of generating a target $y$ is conditioned on the context $C$, which includes $k$ examples and the source $x$. Therefore, the probability can be expressed as:
$$
p_{\mathrm{LLM}}(y \mid C, x)=\prod_{t=1}^T p\left(y_t \mid C, x, y_{<t}\right)
$$
where LLM denotes the parameters of the large language model, and $C=\left\{x_1, y_1, x_2, y_2, \ldots, x_k, y_k\right\}$ is a context string concatenating $k$ training instances. For example, $(x_k, y_k)$ is concatenated with the special character, e.g., ``\verb|\n|'' or ``Sentence: $x$; Sentiment $y$.'' which is denoted as $p_{k}$.
In this paper, we have different example sets $C$ at different stages, $C_{1}$ in the first stage, and $C_{2}$ in the second stage, where $C_{2}$ is a subset of $C_{1}$.


\citet{dai2022can} explains language models as meta-optimizers and understands ICL as a kind of implicit finetuning:
\begin{equation}
    \small
    \begin{aligned}
        \widetilde{\mathcal{F}}_{\text{ICL}}(\mathbf{q})
        & =  W_{\text{ZSL}} \mathbf{q} + \sum_i \left( W_{V} \mathbf{x}^{\prime}_i \otimes \left( W_{K} \mathbf{x}^{\prime}_i \right)^T\right) \mathbf{q} \\
        = & W_{\text{ZSL}} \mathbf{q} + \Delta W_{\text{ICL}} \mathbf{q} \\
        = & \left( W_{\text{ZSL}} + \Delta W_{\text{ICL}} \right) \mathbf{q},
    \end{aligned}
    \label{equ:icl_opti_dual}
\end{equation}
where ZSL denotes the zero-shot learning, which only contains the source $x$; $\mathbf{x} \in \mathbb{R}^{d}$ is the input representation of a query token $t$, and $\mathbf{q} = W_{Q} \mathbf{x} \in \mathbb{R}^{d^{\prime}}$ is the attention query vector;  $\mathbf{x}^{\prime} \in \mathbb{R}^{d}$ denotes the input representations of the example's token; $W_{Q}, W_{K}, W_{V} \in \mathbb{R}^{d^{\prime} \times d}$ are the projection matrices for computing the attention queries, keys, and values, respectively.~\citet{dai2022can} regards $W_{V} X^{\prime}$ as some meta-gradients, which are used to compute the updated matrix $\Delta W_{\text{ICL}}$.

\subsection{Dataset Pruning}
Investigating the data redundant problem not only helps to improve the training efficiency but also helps us understand the representation ability of small data and how many training samples are required and sufficient for a learning system.~\citep{yang2022dataset} proposed to use the Influence Function to accurately and fast estimate the parameter change caused by weighting an example $p$ for the training dataset. The influence of weighting $p$ on the parameters is given by: 

\begin{equation}\label{eq:influence_func}
    \mathcal{I}_{\mathrm{param}}(p) = \frac{\mathrm{d}\hat{\theta}_{\delta,p}}{\mathrm{d}\delta}\bigg|_{\delta=0} = -H_{\hat{\theta}}^{-1}\nabla_{\theta}L(p,\hat{\theta}),
\end{equation}
where $H_{\hat{\theta}} = \frac{1}{n}\sum_{p_i \in \mathcal{D}}\nabla^{2}_{\theta}L(p_i,\hat{\theta})$ is the Hessian and positive definite by assumption, $\mathcal{I}_{\mathrm{param}}(p) \in \mathbb{R}^N$, $N$ is the number of network parameters, $\mathcal{D}$ is the original dataset. After getting the weighting of each example $p$, ~\citep{yang2022dataset} propose generalization-guaranteed pruning or cardinality-guaranteed pruning to get the final compressed dataset $\mathcal{\hat{D}}$.

    
\section{Method}

\subsection{Recall}

Given the training dataset \(\mathcal{D}\) and the query source \(x\), we use BM25~\cite{robertson2009probabilistic} to retrieve a set of relevant examples \(C_1\) for \(x\). For each example \(p_j\) in \(\mathcal{D}\), the BM25 score, denoted as \(R(p_j, x)\), is computed. This score reflects the relevance of example \(p_j\) to the query \(x\). Specifically:

\begin{equation}
R_{j} = \text{BM25}(p_j, x), 
\end{equation}
where \(R_{j}\) is the relevance score of example \(p_j\) with respect to query \(x\).

Subsequently, we form the set \(C_1\) which consists of the top-N examples with the highest relevance scores:

\begin{equation}
C_1 = \{ p_j | j = 1, 2, \ldots, N \},
\end{equation} where \(N\) is the number of examples we wish to recall for the given query \(x\).

\subsection{Influence-Awared Rerank}
For each $p$ in $C_{1}$, we calculate the input representation of tokens $p$ as $P$ and the meta-gradient $G_{p}=W_{V}P$.
To compute $\mathcal{I}_{\mathrm{param}}(p)$ in Eq.~(\ref{eq:influence_func}), we require the Hessian of $p$ for the parameter $W_{V}$, which necessitates the computation of second-order derivatives. However, we only have access to first-order derivatives approximations of the parameters.  Considering that LLMs typically employ cross-entropy loss and maximum likelihood estimation (MLE) for fine-tuning, we have opted to employ the Fisher matrix as an approximation of the Hessian~\cite{barshan2020relatif}. The key to the approximation process is as follows:
\begin{equation}\label{eq:appro_hession}
\nabla^2 f(\mathbf{x}) \approx \nabla f(\mathbf{x}) \nabla f(\mathbf{x})^{\top}
\end{equation}

Then, combining the Eq.~(\ref{eq:influence_func}) with  Eq.~(\ref{eq:appro_hession}), the expression of the influence function for $p$ is:
\begin{equation}\label{eq:influence_fun_vw}
    \mathcal{I}_{\mathrm{param}}(p) =  -\hat{H}_{\hat{\theta}}^{-1}G_{p},
\end{equation}
where $H_{\hat{\theta}} = \frac{1}{n}\sum_{p_i \in \mathcal{D}}G_{p}G_{p}^{\top}$. 

The score of $C_{1}$ is determined by a combination of the influence score and the relevance score, represented as: \begin{equation*}
    \mathcal{S}=\left\{ \lVert\mathcal{I}_{\mathrm{param}}(p_1)\rVert_{F}^{2}+R_{1}, \ldots, \lVert\mathcal{I}_{\mathrm{param}}(p_N)\rVert_{F}^{2}+R_{N} \right\}.
\end{equation*}

Finally, the $K$ in-context learning examples in $C_2$ are chosen by:
\begin{equation*}
     C_{2} = \left\{p_{i}|i \in I \right\}, \text{where}~I = \text{arg}\max_{\substack{I \subseteq \{1,2,\ldots,|{\mathcal{S}}|\} \\ |I| = K}} \mathcal{S}.
\end{equation*}

\begin{table*}[t]

\resizebox{\linewidth}{!}{
\renewcommand{\arraystretch}{1.5}
\begin{tabular}{lc|cc|cc|cc|cc}
\hline
$K=3$                &      & \multicolumn{2}{c}{GPT2-XL}        & \multicolumn{2}{c}{GPT2-Large} & \multicolumn{2}{c}{GPT2-Small}       & \multicolumn{2}{c}{GPT2-Medium}          \\ \cline{3-10}
                              &      & Acc (\%)           & $F_{1}$  (\%)           & Acc (\%)           & $F_{1}$ (\%)            & Acc (\%)           & $F_{1}$  (\%)           & Acc  (\%)          & $F_{1}$ (\%)          \\ \cline{1-10}
Sick                          & BM25 & 42.63          & 33.01          & 27.68          & 26.72          & 31.72          & 23.34          & 31.52          & 26.45           \\  
                              \rowcolor{gray!20} & Ours  & \textbf{47.07} & \textbf{35.28} & \textbf{31.11} & \textbf{28.51} & \textbf{35.35} & \textbf{26.25} & \textbf{32.53} & \textbf{27.16} \\
                              \cline{1-10}
Cola                          & BM25 & 61.84          & 54.83          & 63.09          & 50.24          & \textbf{65.96}          & \textbf{48.79}          & 60.98          & 50.04           \\ 
                             \rowcolor{gray!20} & Ours  & \textbf{63.09} & \textbf{55.53} & \textbf{64.24} & \textbf{50.74} & 65.87          & 48.39          & \textbf{63.95} & \textbf{52.80}    \\
                             \cline{1-10}
Ethos-disability                    & BM25 & 77.01          & 57.44          & 82.76          & 62.42          & 68.97          & 56.92          & 74.71          & 50.26           \\
                             \rowcolor{gray!20} & Ours & \textbf{83.91} & \textbf{66.17} & \textbf{87.36} & \textbf{64.14} & \textbf{74.71} & \textbf{62.96} & \textbf{77.01} & \textbf{51.67}  \\ 
                              \cline{1-10}
Tweet\_eval\_stance\_feminist & BM25 & \textbf{50.75}          & \textbf{46.19}          & 44.78          & 40.96          & 41.79          & 31.64          & 44.78          & \textbf{41.33}          \\
                             \rowcolor{gray!20}  
                             & Ours  & \textbf{50.75}          & 43.27          & \textbf{46.27} & \textbf{41.88} & \textbf{43.28} & \textbf{32.01} & \textbf{46.27} & 38.24        \\ \cline{1-10}
Tweet\_eval\_stance\_hillary  & BM25 & 49.28          & 40.63          & 42.03          & 41.12          & 42.03          & 41.12          & 46.38          & \textbf{44.95}           \\
                              \rowcolor{gray!20} & Ours  & \textbf{53.62} & \textbf{40.68} & \textbf{53.62} & \textbf{51.45} & \textbf{46.38} & 39.24          & \textbf{50.72} & 44.73           \\ 
\cline{1-10}
\textbf{All dataset Avg}  & BM25 & 56.30          & 46.42          & 52.07          & 44.29          & 50.09          & 40.36          & 51.67          & 42.61               \\
                              \rowcolor{gray!20} & Ours  & \textbf{59.69} & \textbf{48.18} & \textbf{56.52} & \textbf{47.34} & \textbf{53.12} & \textbf{41.77} & \textbf{54.10} & \textbf{42.92}         \\                     
\hline

\end{tabular}
}
\caption{Results of four ICL examples. The boldface represents the
best performance.}
\label{tab:k=3}

\end{table*}

\begin{table*}[t]

\resizebox{\linewidth}{!}{
\renewcommand{\arraystretch}{1.5}
\begin{tabular}{lc|cc|cc|cc|cc}
\hline
$K=4$                &      & \multicolumn{2}{c}{GPT2-XL}        & \multicolumn{2}{c}{GPT2-Large} & \multicolumn{2}{c}{GPT2-Small}       & \multicolumn{2}{c}{GPT2-Medium}          \\ \cline{3-10}
                              &      & Acc (\%)            & $F_{1}$ (\%)             & Acc (\%)           & $F_{1}$ (\%)            & Acc (\%)           & $F_{1}$ (\%)             & Acc (\%)           & $F_{1}$ (\%)            \\ \cline{1-10}
Sick                          & BM25                 & 42.83                    & 35.60                   & 31.31          & 31.09          & 33.54          & 26.03          & 31.92          & 28.07         \\
                              \rowcolor{gray!20} & Ours                 & \textbf{46.67}           & \textbf{36.70}          & \textbf{32.53} & \textbf{30.74} & \textbf{39.19} & \textbf{30.42} & \textbf{35.56} & \textbf{29.80}  \\
                                                            \cline{1-10}
Cola                          & BM25                 & 60.98                    & \textbf{54.48}                   & 62.80          & 51.17          & 65.10          & 48.26          & 60.21          & 49.85           \\
        \rowcolor{gray!20} & Ours                  & \textbf{61.36}           & 54.38          & \textbf{63.47} & \textbf{51.51} & \textbf{67.31} & \textbf{48.71} & \textbf{64.91} & \textbf{54.05}   \\
                                      \cline{1-10}
Ethos-disability                    & BM25                 & 81.61                    & 61.33                   & 85.06          & \textbf{61.60}          & 68.97          & 56.92          & 77.01          & 54.78            \\
                              \rowcolor{gray!20} & Ours                  & \textbf{85.06}           & \textbf{64.80}          & \textbf{87.36} & 59.87 & \textbf{72.41} & \textbf{59.60} & \textbf{79.31} & \textbf{56.50} \\
                              \cline{1-10}
Tweet\_eval\_stance\_feminist & BM25                 & 46.27                    & 43.54                   & 43.28          & 36.81          & 43.28          & \textbf{40.64}          & 38.81          & 31.02             \\
                              \rowcolor{gray!20} & Ours                  & \textbf{53.73}           & \textbf{47.36}          & \textbf{47.76} & \textbf{44.47} & \textbf{46.27} & 35.10 & \textbf{44.78} & \textbf{37.11}   \\
                              \cline{1-10}
Tweet\_eval\_stance\_hillary  & BM25                 & 47.83                    & 40.64                   & 34.78          & 34.36          & 39.13          & 33.92          & 40.58          & 39.71             \\
                              \rowcolor{gray!20} & Ours                 & \textbf{47.83}           & \textbf{38.89}          & \textbf{46.38} & \textbf{46.01} & \textbf{46.38} & \textbf{34.81} & \textbf{47.83} & \textbf{44.07}   \\ \cline{1-10}
\textbf{All dataset Avg}  & BM25 & 55.90          & 47.12          & 51.45          & 43.01          & 50.00          & 41.16          & 49.71          & 40.69           \\
                              \rowcolor{gray!20} & Ours  & \textbf{58.93} & \textbf{48.42} & \textbf{55.50} & \textbf{46.52} & \textbf{54.31} & \textbf{41.73} & \textbf{54.48} & \textbf{44.30}          \\  

\hline
\end{tabular}
}
\caption{Results of four ICL examples. The boldface represents the
best performance.}
\label{tab:k=4}

\end{table*}

\section{Experiments}
In this section, we empirically verify the efficiency of our approach. The source code and all experiments have been shared at~\url{https://anonymous.4open.science/r/ICL-F302}.

\subsection{Experiments setup}
This section introduces the detailed information about our experiments.

\textbf{Models.} 
We use the open source GPT2 model family~\cite{radford2019language} (i.e., GPT2-Small, GPT2-Medium, GPT2-Large, GPT2-XL) as a representative of large models to verify the effectiveness of our method. 




\textbf{Datasets.}
We use five datasets spanning four tasks: linguistic analysis, hate speech detection, tweet classification, and semantic similarity. Specifically, we employ \textit{Linguistic Acceptability dataset (Cola)}~\cite{warstadt2019neural}, \textit{online hate speech detection dataset (Ethos and Ethos-disability)}~\cite{mollas2022ethos}, \textit{Tweet\_eval-stance\_feminist} and \textit{Tweet\_eval\_stance\_hillary}~\cite{barbieri2020tweeteval} from Twitter, and \textit{Sentences Involving Compositional Knowledge dataset (Sick)}~\cite{marelli-etal-2014-sick}. 
We use Accuracy and $F_{1}$ score as evaluation metrics. 
Detailed dataset statistics and the prompt templates used can be found in Appendix~\ref{app:temp} and Appendix~\ref{app:dataset}.


\textbf{Implementation Details.}
In the study, we chose $K$ = 3 and $K$ = 4 demonstrations to contrast example selection methods from training data. We set $N$ = 100 for all models.  Sentences were either truncated or supplemented to have a uniform length at 50\% of the average sentence length. Although using multiple transformer layers' meta-gradient might be beneficial, considering the time efficiency, we used the first layer and obtained higher accuracy than baseline models. 

\textbf{Baselines.}
Considering the model proposed in this paper is unsupervised and requires no training, it possesses a higher generalizability and operational efficiency compared to models that undergo supervised training. To ensure a fair comparison, our primary baseline is the unsupervised BM25-based In-Context Example Selection. Previous work~\cite{wang2023learning,gupta2023coverage} has demonstrated that BM25 constitutes a robust baseline for demonstration selection, hence we juxtapose our methodology against BM25. The demonstrations selected by the BM25 are utilized across all GPT2 models.

\subsection{Overall Performance}

Tables \ref{tab:k=3} and \ref{tab:k=4} display results for three and four ICL examples, respectively. Observing the last two rows, our method consistently outperforms across all models and datasets. Using three and four examples, we surpass BM25 by averages of 5.17\% and 6.64\% in all metrics. Specifically, accuracy sees improvements of 6.33\% and 7.80\% over BM25. This underscores our approach's superiority. We found some higher model performance with three ICL examples compared to four, which can be explained by overfitting and example quality. Overfitting in few-shot learning means too many examples leads to adaptation to specific instances rather than general patterns, reducing accuracy on unseen data. Furthermore, if the additional fourth example is of lower quality or relevance, it can degrade model performance.

\section{Related Work}

\subsection{In-context Learning}

In-context learning (ICL) has emerged as a fresh approach in natural language processing (NLP), where large models predict based solely on contexts supplemented by several examples~\cite{dong2022survey,shin2022effect,zhang2023trained,bai2023transformers}. Numerous studies have sought to modify, improve, and comprehend ICL, encompassing topics like prompt tuning~\cite{kim2022self,wang2022iteratively,mishra2022cross}, intrinsic mechanism analysis~\cite{chan2022data,LiIPO23,garg2022can}, evaluations~\cite{srivastava2023beyond,WangMAKMNADASPK22}, and its use across various fields~\cite{sun2023short}, among others.

\subsection{Demonstration Selection}
The goal of demonstration selection is to identify optimal examples for ICL. \citep{liu-etal-2022-makes} demonstrated that choosing the nearest neighbors as in-context examples is an effective approach. The used distance measures include the pre-set L2 distance or the cosine similarity based on sentence embeddings. They introduced KATE, an unsupervised kNN retriever for in-context example selection. \citep{rubin-etal-2022-learning} suggested a two-phase retrieval process for demonstration selection. For a given input, it initially employs an unsupervised retriever (like BM25) to retrieve similar candidate examples and then uses a supervised retriever, EPR, to pick demonstrations from these candidates.
Recent studies indicate that LLMs exhibit strong sensitivity to the examples chosen, resulting in significant performance variations~\cite{nie2022improving}, dependency on example sequence~\cite{lu-etal-2022-fantastically}, and at times, an insensitivity to the actual labels~\cite{min-etal-2022-rethinking}.
Our research focuses on reducing training overhead and condensing crucial data from the training set into in-context examples, which in turn amplifies the ICL's effectiveness. 


\section{Conclusion}
In summary, inspired by LLMs and ICL's potential, we devised a two-stage method using the influence function for optimal in-context example selection. Our approach ensures relevance with the query source and efficiently determines influence scores. The result is an enhancement in ICL performance, with our experiments validating our model's effectiveness. Our framework stands out due to its data-independent nature and minimal reliance on model parameters.

\section{Limitation and Future Work}
 Given resource constraints and page limitations, we provide limited validation in this paper. Our model is a model-agnostic and free-training approach that can be applied to various in-context learning selection models. In the future, we will validate the effectiveness of our model on more large-scale language models, baselines, and datasets.

\bibliography{custom}

\appendix

\section{APPENDIX FOR REPRODUCIBILITY}
\label{sec:appendix}

\subsection{Dataset}
\label{app:dataset}
In our experiments, we employ \textit{Linguistic Acceptability dataset (Cola)}~\cite{warstadt2019neural}, \textit{online hate speech detection dataset (Ethos and Ethos-disability)}~\cite{mollas2022ethos}, \textit{Tweet\_eval-stance\_feminist} and \textit{Tweet\_eval\_stance\_hillary}~\cite{barbieri2020tweeteval} from Twitter, and \textit{Sentences Involving Compositional Knowledge dataset (Sick)}~\cite{marelli-etal-2014-sick}.
Table~\ref{tab:data_analysis} shows more detailed statics of the datasets.
\begin{table}[t]

\small
    \centering

      \resizebox{\linewidth}{!}{
    \begin{tabular}{lccc}
    \toprule
        Dataset& Training Size & Valid Size& Label class  \\ \midrule
Sick& 4439& 495&3\\
Cola& 8551& 1043 &2\\
Ethos-disability& 346& 87&2\\
Tweet\_eval-stance\_feminist& 597& 67&3\\
Tweet\_eval-stance\_hillaryt& 620& 69&3\\
        \bottomrule
    \end{tabular}
  }
      \caption{Dataset statistics.}
    \label{tab:data_analysis}
      
\end{table}

\subsection{Templates.}
\label{app:temp}
Based on \citep{min-etal-2022-rethinking,yoo2022ground}, we employ minimal templates. For the GPT-2 series model, we distinguish between input and label using a space, and this is also applied between demonstration examples. Table \ref{tab:template} presents templates from five datasets, where blue represents the fixed section, and black varies depending on the example content.

\begin{table*}[t]
\small
    \centering

      \resizebox{\linewidth}{!}{
      \renewcommand{\arraystretch}{1.5}
    \begin{tabular}{lll}
    \toprule
        Dataset& example &label \\ \midrule

Sick& [S1] {\color{blue!50} The question is:}[S2]  {\color{blue!50} True or False? Answer:}[Label] &True,Not sure,False\\
Cola&{\color{blue!50} Review:} [S1] {\color{blue!50} Acceptability:}[Label] &unacceptable,acceptable \\
Ethos-disability&{\color{blue!50}  There is a hate based on disability in} [S1]{\color{blue!50} True or False? Answer:} [Label]& False,True\\
Tweet\_eval-stance\_feminist& {\color{blue!50} Tweet:} [S1]{\color{blue!50} Sentiment:}  [Label]& none,against,favor\\
Tweet\_eval-stance\_hillary&{\color{blue!50} Tweet:} [S1]{\color{blue!50} Sentiment:}  [Label]& none,against,favor\\
        \bottomrule
    \end{tabular}
  }
      \caption{Template examples.}
\label{tab:template}
      
\end{table*}

\end{document}